\documentclass[runningheads]{llncs}

\usepackage[T1]{fontenc}
\usepackage{booktabs}
\usepackage{amsmath}
\usepackage{graphicx}
\usepackage{multirow}
\usepackage{xcolor}
\usepackage{float}
\usepackage[section]{placeins}
\usepackage{bbding}

\begin{document}

\title{Extreme Volatility Warning under Label Scarcity via Multi-Source Anomaly Fusion}
\titlerunning{Extreme Volatility Warning via Multi-Source Anomaly Fusion}

\author{Jin Qian\inst{1}\textsuperscript{*}\,\Envelope \and
Zhangzhi Xiong\inst{1}\textsuperscript{*} \and
Mingrui Li\inst{1} \and
Zhen Liu\inst{1}}
\authorrunning{J. Qian et al.}

\institute{ShanghaiTech University, Shanghai, China\\
\email{\{qianjin2023,xiongzhzh2023,limr2023,liuzhen2023\}@shanghaitech.edu.cn}}

\maketitle

\begingroup
\renewcommand{\thefootnote}{*}
\footnotetext{Jin Qian and Zhangzhi Xiong contributed equally to this work.}
\endgroup

\begin{abstract}

Early warning of extreme market volatility is central to financial risk management, but actionable events are rare, nonstationary, and often triggered by exogenous information shocks. In our CSI~300 setting, only $\sim$80 positive samples are observed across 791 training days, making heavily supervised multi-source models unstable. We first analyze a 100K-parameter hierarchical text-signal fusion model (HTSF) and find that added parameterization hurts in this low-label regime. Motivated by this failure, we propose \textbf{AAMSF} (Anomaly-Augmented Multi-Signal Fusion), a semisupervised framework that combines Isolation Forest anomaly scores over market indicators, GDELT events, Chinese financial news, and English media with lightweight Ridge score fusion. We further introduce \textbf{T-AAMSF}, a temporal extension for multi-day anomaly accumulation. On CSI~300 (2018--2023), AAMSF achieves test AUC-ROC \textbf{0.680}, outperforming the strongest unsupervised baseline (0.630) and neural baseline (0.588), while T-AAMSF improves PR-AUC to 0.291. Ablations reveal strong source asymmetry: GDELT and domestic financial news provide complementary risk signals, whereas English media consistently reduces performance, and learned weighting is unreliable under validation noise. These results suggest an empirical design principle for label-scarce financial risk warning: robust anomaly geometry and source reliability can matter more than supervised representation capacity.


\end{abstract}

% ============================================================
\section{Introduction}
% ============================================================

The Chinese capital market, where retail investors account for approximately 80\% of trading volume~\cite{carpenter2021chinese}, is sensitive to herding behavior and policy-driven volatility, making early warning of extreme events a high-stakes risk management problem. Warning signals are distributed across market variables, domestic financial news, structured geopolitical events, and global media coverage. Yet price-based models such as GARCH and extreme value theory cannot exploit these exogenous signals, while existing text-fusion methods mainly target routine return prediction and often treat heterogeneous sources as a monolithic input.

We study a stricter regime: CSI~300 extreme-volatility warning with only $\sim$80 positive samples across 791 training days. Our initial hierarchical text-signal fusion model (HTSF), with source-level recurrent encoders and attention, exposed the core failure mode: increasing architectural complexity degrades test performance under severe label scarcity. In this regime, additional learnable parameters often become liabilities rather than flexibility, because sparse positive supervision amplifies source-specific noise and validation artifacts. This motivates an empirical design principle: preserve source structure, but shift the burden from supervised representation learning to semisupervised anomaly geometry and minimal score-level fusion.

We propose \textbf{AAMSF}, a lightweight anomaly-based framework combining per-source Isolation Forest scores with a 16-parameter Ridge component for market features, and \textbf{T-AAMSF}, a temporal extension that accumulates anomaly evidence across consecutive days. Our contributions are: \textbf{(1)} a failure analysis showing that a 100K-parameter HTSF model and its lighter variants fail under rare-event label scarcity; \textbf{(2)} a lightweight anomaly-fusion framework achieving test AUC-ROC 0.680; \textbf{(3)} temporal anomaly accumulation reaching the best PR-AUC (0.291); and \textbf{(4)} source and weighting analysis showing that GDELT and Chinese financial news are complementary, English news is actively harmful in this China-market setting, and equal weighting can dominate learned weighting under validation noise.

% ============================================================
\section{Related Work}
% ============================================================
\textbf{Financial text mining.}
Tetlock~\cite{tetlock2007giving} links media pessimism to stock returns, and Bollen et al.~\cite{bollen2011twitter} find Twitter mood predictive of market movements.
FinBERT~\cite{araci2019finbert} adapts BERT to English financial text, LaBSE~\cite{feng2022labse} aligns 109 languages, and Hu et al.~\cite{hu2018listening} show social media is predictive in Chinese markets.
Recent financial LLMs and generative AI systems further expand the modeling capacity available for financial NLP, but they do not remove the day-level label scarcity and source-reliability issues faced by rare-event warning.
Most existing work targets routine returns rather than rare extreme events, and treats sources uniformly rather than as separately curated signals.

\textbf{Extreme event prediction.}
Tail-risk methods include GARCH~\cite{bollerslev1986generalized} and EVT~\cite{mcneil2000estimation}. Jump-diffusion models~\cite{kou2002jump} provide another route, while Zhang et al.~\cite{zhang2022tail} apply gradient boosting to the Chinese market.
All rely on price dynamics alone and ignore the exogenous textual and event signals surrounding extreme events.

\textbf{Classical ML vs.\ deep learning in finance.}
When samples are limited and signals noisy, classical ML often beats deep learning in finance: random forests outperform neural nets in stock direction~\cite{ballings2015evaluating}, shallow learners beat deeper models in asset pricing~\cite{gu2020empirical}, and ARIMA can outperform deep forecasters in stock and FX prediction~\cite{shah2021forecasting}.
Deep models dominate only with very large datasets, e.g., billions of order-book observations~\cite{sirignano2019universal}.
Our setting (hundreds of trading days, tens of positives) sits squarely in this low-sample regime, motivating AAMSF's near-parameter-free design.

\textbf{Multi-source fusion.}
Attention-based fusion dominates heterogeneous financial signal combination.
Chen et al.~\cite{chen2022hierarchical} use hierarchical attention over multi-view news, and Zhao et al.~\cite{zhao2023structured} apply structured multi-head attention on $\sim$850 trading days---comparable to our scale.
Both rely on substantial learnable parameters and have not been tested under extreme label scarcity, a regime probed by our HTSF analysis.

\textbf{Anomaly detection for finance.}
Anomaly detection flags distributional outliers without labels, making it attractive for rare-event settings.
Isolation Forest~\cite{liu2008isolation} uses random partitioning path lengths; Aggarwal and Sathe~\cite{aggarwal2015outlier} argue equal weighting beats learned weighting when detector quality is unreliable.
However, it is rarely used as the primary mechanism for extreme-event early warning under severe label scarcity.

\textbf{Positioning.}
Existing work either pursues expressive supervised prediction or anomaly detection without source fusion.
We study a distinct regime characterized by severe label scarcity, heterogeneous signals, and rare-event forecasting, where minimizing learnable parameters becomes central.

% ============================================================
\section{Problem Statement}
% ============================================================

\begin{figure}[H]
    \centering
    \includegraphics[width=\columnwidth]{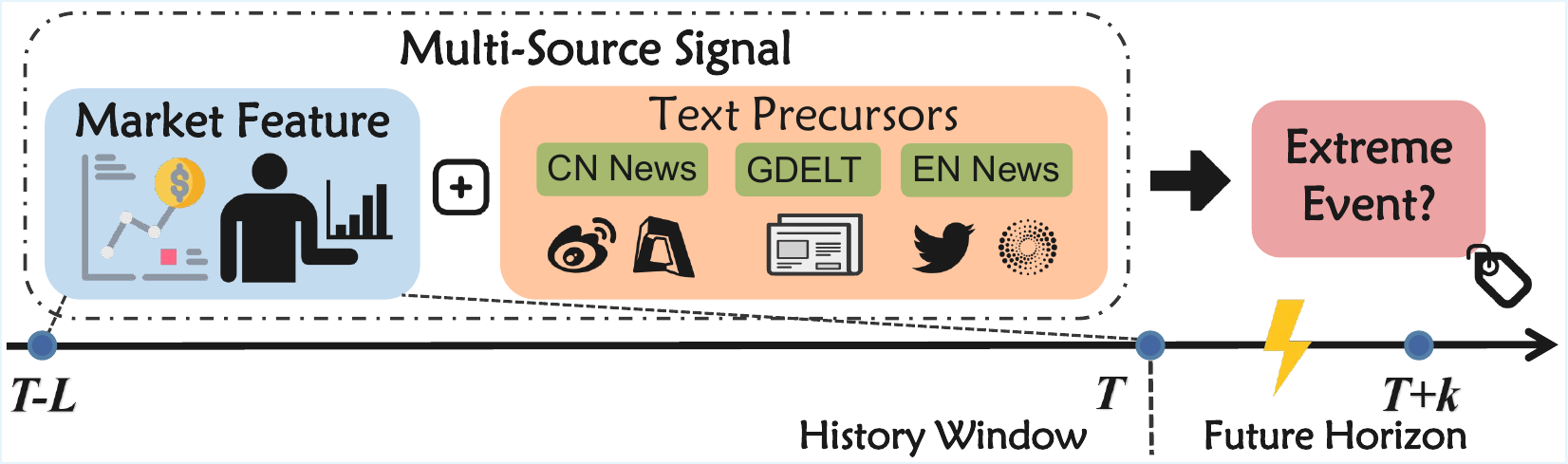}
    \caption{Illustration of Extreme Event Task}
    \label{fig:my_pdf_image}
\end{figure}

\textbf{Task.}
As illustrated in Figure~\ref{fig:my_pdf_image}, given a market feature vector $\mathbf{m}_t$ and per-source text feature vectors $\mathbf{x}^{(s)}_t$ for $s{=}1,\ldots,S$ on day $t$, predict whether an extreme volatility event occurs within $[t{+}1, t{+}3]$:
\begin{equation}
f\!\left(\mathbf{m}_t, \mathbf{x}^{(1)}_t, \ldots, \mathbf{x}^{(S)}_t\right) \to y_t \in \{0,1\}.
\end{equation}

\textbf{Label (EV3).}
Let $r_t$ be the daily log return and let $\mu_{60}, \sigma_{60}$ denote the 60-day rolling mean and std. of $|r_t|$; the z-score is $z_t = (|r_t|-\mu_{60})/\sigma_{60}$. Day $t$ is labelled an \emph{Extreme Volatility within 3 days} (EV3) event if $y_t = \mathbf{1}[\max_{i=1}^{3} z_{t+i} > \theta_{96.5}]{=}1$, where $\theta_{96.5}$ is the 96.5th percentile of training z-scores---a standard tail threshold that gives about 10\% positives on train and 14\% on the held-out splits.

\textbf{Data split.}
Strictly temporal over 2018--2023: Train ($\le$2021-06, 791 days, 80 positives), Validation (2021-07 to 2022-06, 242 days, 33 positives), Test ($\ge$2022-07, 366 days, 50 positives).

AAMSF computes per-day IF scores independently, while T-AAMSF (\S\ref{sec:taamsf}) aggregates them across a fixed-length window via temporal fusion.

% ============================================================
\section{Data Collection and Preprocessing}
% ============================================================
HTSF originally represented each trading day by article-level LaBSE embeddings compressed to 64 dimensions via PCA, organized into per-source tensors alongside standardized market features---roughly $32{,}000$ floats per sample. Combined with only 80 positive training days, this representation proved prone to overfitting, motivating AAMSF's switch to compact daily aggregates over four heterogeneous information streams (Figure~\ref{fig:preprocessing}):

\begin{figure}[H]
    \centering    \includegraphics[width=0.5\textwidth]{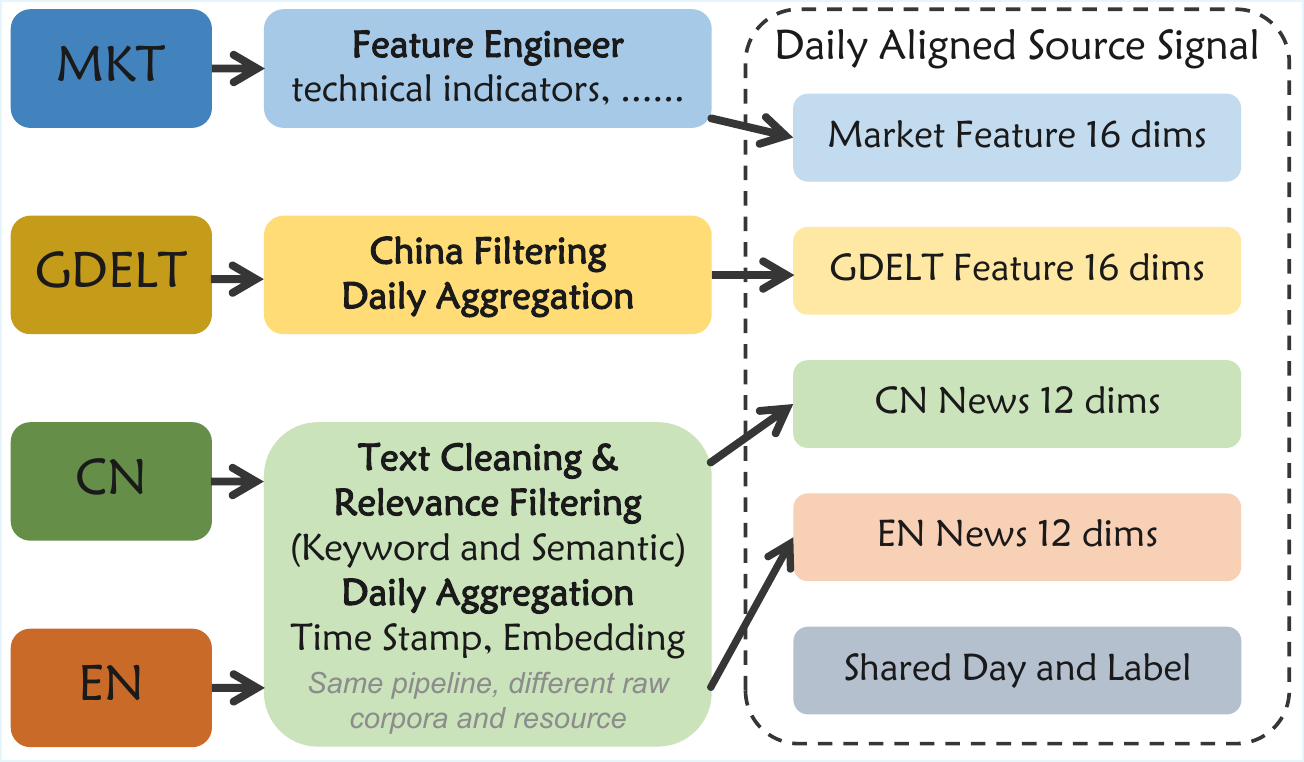} 
    \caption{Illustration of Data Preprocessing}
    \label{fig:preprocessing}
\end{figure}

\textbf{Market data} (16 features): CSI~300 daily OHLCV from AKShare; derived features include log returns, realized volatility (5/10/20-day), volume ratios, technical indicators (RSI, MACD, Bollinger Bands, moving-average crossovers), and microstructure measures (intraday range, overnight gap, Amihud illiquidity).

\textbf{GDELT} (16 features): China-filtered events from the GDELT Project~\cite{leetaru2013gdelt}; daily aggregates of average tone, Goldstein conflict--cooperation scale, quad-class conflict/cooperation ratios, and event-count statistics.

\textbf{Chinese news} (12 features): 17{,}619 articles from domestic Chinese financial media (including CCTV), filtered by financial keywords and LaBSE~\cite{feng2022labse} semantic similarity ($\ge$0.35). Daily features include sentiment statistics (SnowNLP~\cite{snownlp}), article-count z-scores, crisis-keyword ratios, and embedding-based novelty and crisis-similarity measures.

\begin{sloppypar}
\textbf{English news} (12 features): 153{,}851 articles from Reuters, NYT, The Guardian, Twitter, and Hacker News, filtered by China-related keywords and the same LaBSE threshold. Daily features mirror the Chinese news pipeline but use VADER~\cite{hutto2014vader} sentiment.
\end{sloppypar}

\textbf{Preprocessing.} Articles published after market close (15:00 CST) are attributed to the next trading day. To make text features amenable to IF's geometric detection, we apply rolling z-score (20-day), rate-of-change, and rolling percentile (60-day) transforms to news features, converting absolute values into ``deviation from recent baseline'' representations; market and GDELT features, which already encode relative quantities, are kept raw.

% ============================================================
% ============================================================
\section{Method}
% ============================================================

\subsection{Supervised Overfitting Analysis: HTSF}

\textbf{HTSF} (Hierarchical Temporal Source Fusion) is a 5-level deep architecture stacking article-level LaBSE encoders with temporal-distance encoding, per-source attention pooling, per-source LSTMs, market-conditioned cross-source attention, and a classification head, following the dominant paradigm in multi-source financial prediction~\cite{chen2022hierarchical,zhao2023structured}.

HTSF has approximately 100K trainable parameters. With only 80 positive training samples, the parameter-to-positive ratio is about 1250:1, highlighting the extreme mismatch between model complexity and the effective signal available for learning. Results in Table~\ref{tab:htsf} show that HTSF performs worse than every simpler deep-learning baseline.

\begin{table}[!htbp]
\small
\centering
\caption{HTSF vs.\ simpler DL baselines}
\vspace{-2mm}
\label{tab:htsf}
\begin{tabular}{@{}lcccc@{}}
\toprule
\textbf{Model} & \textbf{Params} & \textbf{AUC} & \textbf{PR-AUC} & \textbf{F1} \\
\midrule
Embedding MLP        & $\sim$10K  & 0.508          & 0.243          & 0.179 \\
Flat LSTM            & $\sim$20K  & 0.468          & 0.238          & 0.039 \\
Transformer Encoder  & $\sim$30K  & \textbf{0.521} & \textbf{0.295} & \textbf{0.215} \\
HTSF                 & $\sim$100K & 0.443          & 0.120          & 0.186 \\
\bottomrule
\end{tabular}
\end{table}

\textbf{Root cause.} The bulk of HTSF's $\sim$70K extra parameters (per-source LSTMs and market-conditioned cross-source attention) yields no learnable signal in this regime; despite its complexity, HTSF underperforms every simpler baseline in Table~\ref{tab:htsf}. To check whether source attribution itself requires learned parameters, we attach a zero-parameter feature-ablation head to the Transformer encoder---a variant we call \textbf{HTSFLite}. Because the head is purely post-hoc, HTSFLite shares the Transformer's architecture and produces identical predictive metrics (Table~\ref{tab:htsf}, Transformer row), while additionally exposing per-source importance scores. This confirms that source attribution can be obtained without any learned parameters, directly motivating AAMSF's design choice: \emph{compute source contribution post-hoc rather than learn it}.

\subsection{Lightweight Anomaly Fusion: AAMSF}

The HTSF analysis motivates \textbf{AAMSF} (Anomaly-Augmented Multi-Signal Fusion), designed around minimizing learnable parameters: it combines unsupervised Isolation Forest~\cite{liu2008isolation} anomaly detection with a lightweight Ridge regressor, fused at the score level.

\textbf{Stage 1: Multi-view IF (0 learnable parameters).}
For each source, we train a 10-seed IF ensemble (200 trees, contamination=0.1) on unlabeled training data. Per-source scores are combined with \emph{fixed equal weights} ($1/N$); the choice is empirically validated against 10 learned weighting strategies in \S\ref{sec:weighting}.

\textbf{Stage 2: Ridge regression (16 parameters).}
A Ridge regressor on 16 market features predicts continuous rv3 (mean 3-day absolute return), so all 791 days inform training---versus only 80 positives in binary classification.

\textbf{Stage 3: Fixed-weight fusion.}
\[
\text{Score} = 0.85\,\text{norm}(\text{IF}_{\text{combined}})
+ 0.15\,\text{norm}(\text{Ridge}).
\]
Here, norm denotes min--max normalization on training statistics. The 85/15 split is fixed by design: validation-tuned weights overfit on the 242-sample validation set (val AUC 0.573, test AUC 0.466, a $-0.107$ gap), whereas the fixed prior gives a stable test AUC of 0.680.

\textbf{Threshold calibration.} The binary decision threshold is calibrated on validation by maximizing F1.

\subsection{Temporal Extension: T-AAMSF}
\label{sec:taamsf}

AAMSF's Stage~1 IF processes each day independently. Yet extreme volatility events often show \emph{signal build-up}: anomaly indicators tend to intensify over 2--3 days before the event. \textbf{T-AAMSF} captures this by modifying Stage~1 alone---replacing each per-source IF score with an exponential-decay weighted average over the past $L$ days:
\begin{equation}
\widetilde{\text{IF}}_s(t) = \sum_{k=0}^{L-1} w_k \cdot \text{IF}_s(t{-}k), \quad \sum_{k=0}^{L-1} w_k = 1.
\end{equation}
The aggregated $\widetilde{\text{IF}}_s$ then enters the same Stage~2--3 pipeline of AAMSF, so AAMSF and T-AAMSF coincide when $L{=}1$. Default weights $\mathbf{w}{=}[0.6,0.3,0.1]$ are fixed priors (zero learnable parameters); we additionally test source-specific decay (slower for GDELT, faster for English news).

\textbf{Results.} Holding the source set fixed at all four signals to isolate the temporal effect, T-AAMSF improves AUC by $+0.052$ over raw-feature AAMSF and $+0.012$ over IF-aware AAMSF (Table~\ref{tab:taamsf}). More importantly, T-AAMSF attains \textbf{PR-AUC 0.291}---higher than AAMSF-Opt's 0.276 on the curated 3-source set---the best in our study. Multi-day anomaly accumulation thus carries signal beyond same-day snapshots. Given the small gap between AAMSF-Opt and T-AAMSF on both metrics (within $\pm$0.015), we adopt AAMSF as the primary backbone for the subsequent ablation and analysis; extending T-AAMSF to the optimal 3-source configuration is an immediate next step.

\begin{table}[!htbp]
\small
\centering
\caption{T-AAMSF vs.\ AAMSF on the 4-source set. AAMSF-Opt (3-source) shown for reference.}
\vspace{-2mm}
\label{tab:taamsf}
\begin{tabular}{@{}lccc@{}}
\toprule
\textbf{Model} & \textbf{AUC} & \textbf{PR-AUC} & \textbf{F1} \\
\midrule
AAMSF (raw, 4-src) & 0.613 & 0.216 & 0.317 \\
AAMSF (mixed, 4-src) & 0.653 & 0.263 & 0.331 \\
\textbf{T-AAMSF} (equal decay) & \textbf{0.665} & \textbf{0.291} & 0.305 \\
T-AAMSF (src-specific decay) & 0.662 & 0.287 & 0.322 \\
\midrule
\emph{AAMSF-Opt (3-src, ref.)} & \emph{0.680} & \emph{0.276} & \emph{0.333} \\
\bottomrule
\end{tabular}
\end{table}

% ============================================================
\section{Experimental Results}
% ============================================================

\subsection{Main Results}

\begin{figure}[!htbp]
    \centering
    \includegraphics[width=\columnwidth]{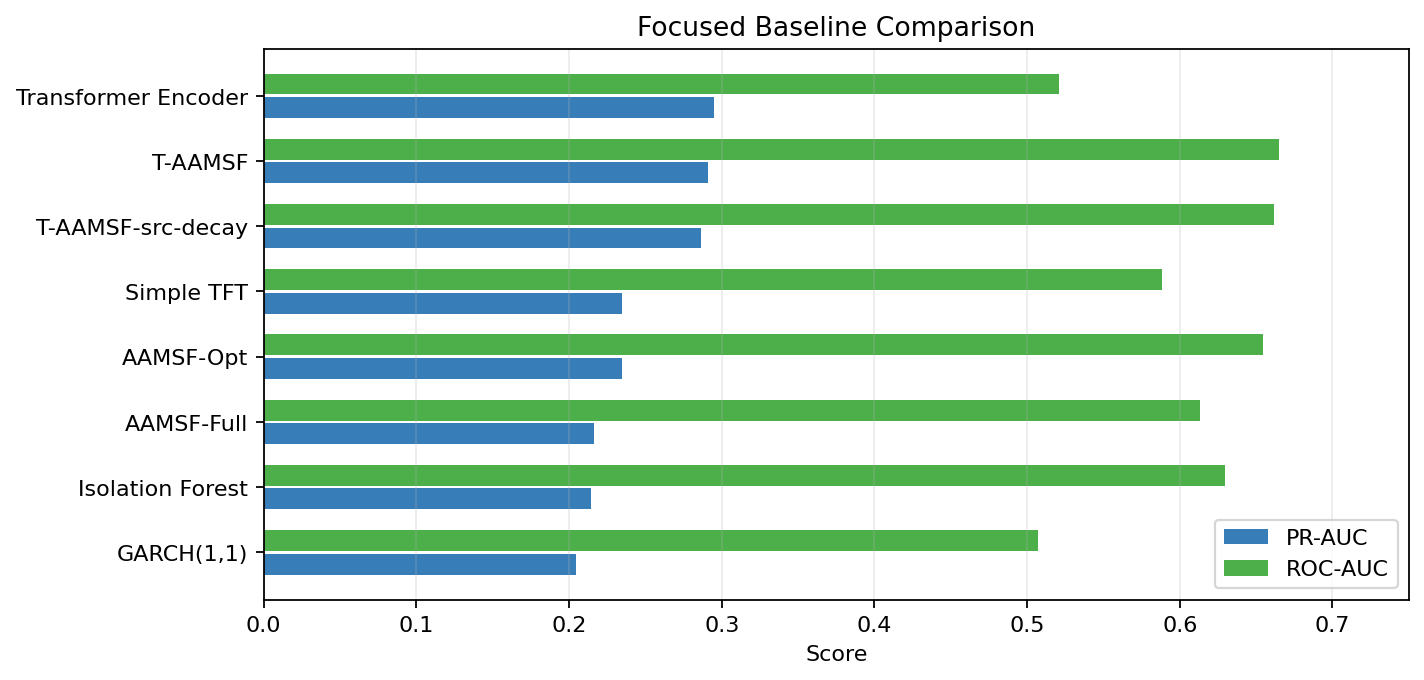}
    \caption{Focused comparison on the CSI~300 test set: representative baselines vs.\ the AAMSF/T-AAMSF family under the 4-source configuration.}
    \label{fig:focused}
\end{figure}

Figure~\ref{fig:focused} visualizes the head-to-head ROC-AUC and PR-AUC of representative baselines against the AAMSF/T-AAMSF family on the 4-source configuration. Three patterns stand out: (i)~AAMSF and T-AAMSF dominate the ROC-AUC axis over the strongest neural (Simple TFT) and unsupervised (Isolation Forest) baselines; (ii)~T-AAMSF takes the lead on PR-AUC, confirming that multi-day anomaly aggregation contributes signal beyond same-day snapshots; (iii)~source-specific decay (T-AAMSF-src-decay) is marginally worse than equal decay, consistent with our broader finding that learned per-component priors are unreliable in this low-sample regime.

Table~\ref{tab:main} additionally reports the curated 3-source variant \textbf{AAMSF-Opt} (Market+GDELT+CN news with mixed IF-aware preprocessing), which achieves AUC-ROC \textbf{0.680}---the best in our study---by excluding the harmful English news (\S\ref{sec:source-quality}). AAMSF-Opt outperforms the best unsupervised baseline (IF: 0.630, $+7.9\%$) and the best neural baseline (Simp.\ TFT: 0.588, $+15.6\%$).

\begin{table}[H]
\small
\centering
\caption{Test set performance of representative baselines.}
\label{tab:main}
\begin{tabular}{@{}lccc@{}}
\toprule
\textbf{Model} & \textbf{AUC} & \textbf{PR-AUC} & \textbf{F1} \\
\midrule
GARCH(1,1)          & 0.507          & 0.205          & 0.038 \\
Isolation Forest    & 0.630          & 0.214          & 0.137 \\
Logistic Regression & 0.469          & 0.122          & 0.248 \\
LightGBM            & 0.305          & 0.096          & 0.031 \\
TF-IDF + LR         & 0.456          & 0.160          & 0.188 \\
Embedding MLP       & 0.508          & 0.242          & 0.179 \\
Flat LSTM           & 0.468          & 0.238          & 0.039 \\
Transformer Encoder & 0.521          & \textbf{0.295} & 0.215 \\
HTSF                & 0.443          & 0.120          & 0.186 \\
Simp.\ TFT          & 0.588          & 0.235          & 0.240 \\
\midrule
\textbf{AAMSF-Opt}  & \textbf{0.680} & 0.276          & \textbf{0.333} \\
\bottomrule
\end{tabular}
\end{table}

\FloatBarrier
\subsection{Case Study: Warning Windows}
\label{sec:case}

To verify that AAMSF-Opt improves more than aggregate ranking metrics, we inspect the held-out test period and group consecutive EV3-positive days into volatility warning windows. We report the two detected positive clusters with the highest AAMSF-Opt peak scores, plus the detected cluster with the largest GDELT-minus-market gap as a source-asymmetry diagnostic. This selection rule avoids relying only on visually convenient examples.

Table~\ref{tab:case_study} shows that AAMSF-Opt captures joint-signal regimes as well as asymmetric source behavior. The October--November 2022 window is a joint-signal case: both market and GDELT anomaly scores are elevated, and AAMSF-Opt fires on five of the eleven positive days. The November 2022 window is more news-driven, where the market-only score stays below 0.20 but GDELT and MKT+CN both exceed 0.43. The December 2023 window is the strongest GDELT-dominant detected case: GDELT reaches 0.589 while market-only remains at 0.154. These examples support the intended role of AAMSF as a lightweight warning mechanism: it can combine market stress with external information, while still allowing one informative source family to dominate when price-only signals are weak.

\begin{table}[!htbp]
\footnotesize
\centering
\caption{Representative held-out volatility warning windows. Hits count AAMSF-Opt positive predictions on EV3-positive days; scores are peak daily scores within each window.}
\vspace{-2mm}
\label{tab:case_study}
\setlength{\tabcolsep}{3pt}
\begin{tabular}{@{}lcccccc@{}}
\toprule
\textbf{Window} & \textbf{Days} & \textbf{Hits} & \textbf{AAMSF} & \textbf{MKT} & \textbf{GDELT} & \textbf{MKT+CN} \\
\midrule
2022-10-19--11-03 & 11 & 5 & 0.477 & 0.551 & 0.618 & 0.431 \\
2022-11-24--11-28 & 3  & 2 & 0.430 & 0.197 & 0.484 & 0.439 \\
2023-12-25--12-27 & 3  & 2 & 0.410 & 0.154 & 0.589 & 0.298 \\
\bottomrule
\end{tabular}
\end{table}

\subsection{Per-Source IF Analysis}
\label{sec:source-quality}

For each source we train a single-source IF ensemble and report test AUC under both raw and IF-aware preprocessing (Table~\ref{tab:src_if}). GDELT and Chinese news are strong (AUC $>$ 0.6); English news falls below random---its anomaly structure is anti-correlated with CSI~300 extremes. The IF-aware transform is source-dependent: it boosts Chinese news (+0.067) but hurts GDELT ($-0.111$), since GDELT already encodes relative quantities (e.g., conflict\_ratio) that rolling z-scoring inflates with noise. AAMSF-Opt thus uses mixed preprocessing: raw GDELT, IF-aware Chinese news, raw market.

\begin{table}[!htbp]
\small
\centering
\caption{Per-source IF AUC: raw vs.\ IF-aware preprocessing.}
\vspace{-2mm}
\label{tab:src_if}
\begin{tabular}{@{}lccc@{}}
\toprule
\textbf{Source} & \textbf{Raw AUC} & \textbf{IF-aware AUC} & \textbf{$\Delta$} \\
\midrule
Market & 0.607 & 0.607 & \phantom{-}0.000 \\
GDELT & \textbf{0.625} & 0.514 & $-$0.111 \\
Chinese news & 0.553 & \textbf{0.620} & \phantom{-}\textbf{+0.067} \\
English news & 0.488 & 0.467 & $-$0.021 \\
\bottomrule
\end{tabular}
\end{table}

\textbf{Validation unreliability.} With only 242 validation days, per-source quality estimates are noisy: Table~\ref{tab:src_valtest} shows GDELT looking worst on validation (0.423, below random) yet strongest on test (0.625)---a $+0.202$ gap. Validation-based weighting would systematically assign too little weight to GDELT, motivating AAMSF's equal-weight design.

\begin{table}[!htbp]
\small
\centering
\caption{Per-source IF AUC: validation vs.\ test (mixed IF-aware).}
\vspace{-2mm}
\label{tab:src_valtest}
\begin{tabular}{@{}lccc@{}}
\toprule
\textbf{Source} & \textbf{Val AUC} & \textbf{Test AUC} & \textbf{Gap} \\
\midrule
Chinese news & \textbf{0.639} & 0.620 & $-$0.019 \\
Market & 0.553 & 0.607 & \phantom{-}+0.054 \\
GDELT & 0.423 & \textbf{0.625} & \textbf{\phantom{-}+0.202} \\
\bottomrule
\end{tabular}
\end{table}

\subsection{Source Ablation}
We evaluate single-source, pairwise, and multi-source configurations under the fixed AAMSF protocol (Table~\ref{tab:ablation}). GDELT is the strongest standalone source (AUC 0.624); adding it to market features gives the largest pairwise gain ($+0.050$). The 3-source set Market+GDELT+CN reaches AUC \textbf{0.680}; adding English news drops it back to 0.613, consistent with English news's anti-correlated anomaly structure (\S\ref{sec:source-quality}).

\begin{table}[!htbp]
\small
\centering
\caption{Source ablation (AAMSF, mixed IF-aware preprocessing)}
\vspace{-2mm}
\label{tab:ablation}
\begin{tabular}{@{}lccc@{}}
\toprule
\textbf{Configuration} & \textbf{AUC} & \textbf{PR-AUC} & \textbf{F1} \\
\midrule
Market only & 0.608 & 0.214 & 0.162 \\
GDELT only & 0.624 & 0.243 & 0.253 \\
Market + GDELT & 0.658 & 0.250 & 0.180 \\
Market + GDELT + CN & \textbf{0.680} & \textbf{0.276} & \textbf{0.333} \\
Market + GDELT + EN & 0.605 & 0.207 & 0.277 \\
All 4 sources & 0.613 & 0.216 & 0.317 \\
\bottomrule
\end{tabular}
\end{table}

\textbf{Leave-one-out.} Removing each source from the optimal 3-source set confirms all three contribute positively (Table~\ref{tab:loo}); GDELT's removal causes the largest drop, consistent with its highest standalone IF AUC.

\begin{table}[!htbp]
\small
\centering
\caption{Leave-one-out on the optimal 3-source AAMSF.}
\vspace{-2mm}
\label{tab:loo}
\begin{tabular}{@{}lcc@{}}
\toprule
\textbf{Removed} & \textbf{Remaining AUC} & \textbf{$\Delta$AUC} \\
\midrule
None (full 3-source) & 0.680 & --- \\
$-$Market & 0.660 & $-$0.020 \\
$-$GDELT & 0.657 & $-$0.023 \\
$-$Chinese news & 0.662 & $-$0.018 \\
\bottomrule
\end{tabular}
\end{table}

\textbf{Within-GDELT filtering.} A natural concern is whether GDELT's value comes solely from its finance-tagged subset. Restricting GDELT to finance-tagged events drops Market+GDELT performance markedly (Table~\ref{tab:gdelt_filter}), suggesting that GDELT's non-financial content---geopolitical, regulatory, and macroeconomic events---carries useful macro risk signal rather than noise.

\begin{table}[!htbp]
\small
\centering
\caption{Within-GDELT robustness: finance-tagged vs.\ all China-related events (Market + GDELT).}
\vspace{-2mm}
\label{tab:gdelt_filter}
\begin{tabular}{@{}lccc@{}}
\toprule
\textbf{GDELT subset} & \textbf{AUC} & \textbf{PR-AUC} & \textbf{F1} \\
\midrule
All China-related (default) & \textbf{0.658} & \textbf{0.250} & 0.180 \\
Finance-tagged only         & 0.583          & 0.197          & \textbf{0.189} \\
\midrule
$\Delta$ (filtered $-$ default) & $-$0.075 & $-$0.053 & \phantom{-}+0.009 \\
\bottomrule
\end{tabular}
\end{table}

\subsection{Source Weighting Strategy Comparison}
\label{sec:weighting}

\begin{sloppypar}
To validate AAMSF's fixed equal-weight choice, we compare 10 alternative weighting strategies on the same 3-source backbone; only the rule combining per-source IF scores varies. The strategies cover parameter-free rules (Equal, Inverse-variance), performance-based rules (AUC-Prop, Softmax-AUC, Bayesian), learned rules (Stacking, Stacking-CV, Attention, MKL), and locally adaptive LSCP.
\end{sloppypar}

\begin{table}[!htbp]
\small
\centering
\caption{Source weighting strategy comparison on AAMSF-Opt. ``Wt.\,GDELT'' = weight assigned to GDELT.}
\vspace{-2mm}
\label{tab:weighting}
\begin{tabular}{@{}llccc@{}}
\toprule
\textbf{Strategy} & \textbf{Category} & \textbf{Wt.\,GDELT} & \textbf{AUC} & \textbf{F1} \\
\midrule
\textbf{Equal} (used in AAMSF) & param-free & 0.333 & \textbf{0.680} & 0.333 \\
Inverse-variance & param-free & 0.348 & 0.679 & \textbf{0.359} \\
Bayesian (Dirichlet) & performance & 0.296 & 0.680 & 0.356 \\
LSCP (local) & local & 0.274 & 0.680 & 0.356 \\
Attention & learned & 0.042 & 0.668 & 0.252 \\
Stacking & learned & 0.000 & 0.663 & 0.263 \\
MKL & learned & 0.000 & 0.658 & 0.245 \\
Softmax-AUC & performance & 0.011 & 0.645 & 0.301 \\
Stacking-CV & learned & 0.000 & 0.608 & 0.162 \\
\bottomrule
\end{tabular}
\end{table}

Test AUC tracks GDELT's assigned weight almost monotonically: strategies allocating $\ge 0.27$ to GDELT (Equal, Inverse-variance, Bayesian, LSCP) reach AUC $\approx 0.680$; learned methods drive GDELT to near zero and drop to $0.61$--$0.67$. The mechanism is exactly the val-test divergence in Table~\ref{tab:src_valtest}: each learned method ``sees'' GDELT below random on validation and prunes it, losing the strongest test-time signal. This empirically confirms Aggarwal \& Sathe~\cite{aggarwal2015outlier}: when detector quality cannot be reliably estimated, equal weighting dominates.

\subsection{Ridge Component Feature Importance}
\label{sec:ridge}

Although the Ridge regressor contributes only 15\% of the fused score, its weights expose which market features drive the supervised signal. Table~\ref{tab:ridge_top} lists the top-10 features by absolute Ridge weight on the rv3 target. Volatility-related features (\texttt{abs\_volatility}, \texttt{BB\_width}, \texttt{realized\_vol\_10d}, \texttt{intraday\_range}) jointly account for $\sim$38\% of total weight, confirming that recent realized volatility is the strongest supervised proxy for forward extreme events.

\begin{table}[!htbp]
\small
\centering
\caption{Top-10 Ridge feature weights (rv3 target, 16-feature market subset).}
\vspace{-2mm}
\label{tab:ridge_top}
\begin{tabular}{@{}lcc|lcc@{}}
\toprule
\textbf{Feature} & \textbf{Wt.} & \textbf{Type} & \textbf{Feature} & \textbf{Wt.} & \textbf{Type} \\
\midrule
abs\_volatility & 15.3\% & vol & BB\_width & 8.0\% & vol \\
MACD\_hist & 12.7\% & tech & intraday\_range & 7.6\% & micro \\
MA\_5\_20\_ratio & 11.8\% & trend & realized\_vol\_10d & 7.0\% & vol \\
z\_score & 9.8\% & ret & RSI\_14 & 5.6\% & tech \\
BB\_pctb & 8.8\% & tech & overnight\_gap & 4.9\% & micro \\
\bottomrule
\end{tabular}
\end{table}

\FloatBarrier

% ============================================================
\section{Conclusion and Discussion}
% ============================================================

We presented \textbf{AAMSF}, a near-parameter-free framework for early warning of extreme volatility events in the Chinese stock market under severe label scarcity. AAMSF couples multi-view Isolation Forest anomaly detection with a 16-parameter Ridge regressor and fixed 85/15 score-level fusion. On CSI~300 (2018--2023, strict temporal split), AAMSF reaches test AUC \textbf{0.680}, outperforming 10 representative baselines spanning statistical, classical supervised, and deep neural paradigms. Source-relevance analysis revealed an asymmetric value structure: GDELT and Chinese news are complementary, while English news is anti-correlated in this setting. A 10-strategy weighting comparison further supports AAMSF's equal-weight design under noisy validation estimates. We also proposed \textbf{T-AAMSF}, a temporal extension capturing multi-day anomaly build-up. Overall, the study supports a practical empirical principle for financial data mining: in rare-event risk warning, anomaly geometry and source reliability can be more robust than supervised representation learning.

\subsection{When Supervision Becomes a Liability}
\label{sec:curse}

Table~\ref{tab:params_pos} arranges representative models by their parameter-to-positive ratio against test AUC. The pattern is unambiguous: model performance degrades roughly monotonically as the ratio grows. AAMSF places virtually all its capacity on the non-parametric IF (ratio $\approx 0$) and a 16-parameter Ridge regressor trained on \emph{all} 791 days via the regression target rv3 (ratio $\approx 0.02$), in stark contrast to deep supervised models that pile thousands of parameters per training positive.

\begin{table}[H]
\small
\centering
\caption{Parameter-to-positive ratio vs.\ test AUC. Ridge uses the regression target rv3 so all 791 days are informative.}
\vspace{-2mm}
\label{tab:params_pos}
\begin{tabular}{@{}lccc@{}}
\toprule
\textbf{Model} & \textbf{Params} & \textbf{Params/Pos} & \textbf{Test AUC} \\
\midrule
IF (AAMSF) & 0 (non-param.) & 0 & 0.625 \\
Ridge (AAMSF) & 16 & 0.02 & 0.559 \\
LightGBM (mkt+txt) & $\sim$18K & $\sim$225 & 0.305 \\
LSTM + Focal & 89K & $\sim$1{,}114 & 0.449 \\
Simplified TFT & 117K & $\sim$1{,}462 & 0.588 \\
HTSF & 100K & $\sim$1{,}250 & 0.443 \\
\bottomrule
\end{tabular}
\end{table}

\subsection{Limitations and Outlook}

\textbf{Cross-market transfer.}
We evaluate only the CSI~300. Its retail-dominated structure and policy sensitivity may differ substantially from other markets. Future work should evaluate whether AAMSF generalizes across indices with different participant compositions and volatility regimes (e.g., S\&P~500, Nikkei~225, Hang Seng), testing whether anomaly-driven fusion remains effective beyond the Chinese market.

\textbf{T-AAMSF on the optimal source set.} Our T-AAMSF results (\S\ref{sec:taamsf}) use the 4-source configuration; running T-AAMSF on the curated 3-source set (the AAMSF-Opt setting) is an immediate next step.

\textbf{Deployment-oriented validation.}
This work evaluates ranking and thresholded warning quality using historical splits. A practical financial early-warning system would additionally require transaction-cost-aware backtesting, calibration under changing volatility regimes, and robustness checks around major policy and macroeconomic event windows.

\begin{credits}
\subsubsection{\discintname}
The authors have no competing interests to declare that are relevant to the content of this article.
\end{credits}

\FloatBarrier

\bibliographystyle{splncs04}
\bibliography{references}

\end{document}